\pdfoutput=1
\documentclass{article}

\usepackage[margin=1in]{geometry}
\usepackage{wrapfig}
\usepackage{multirow}
\usepackage{adjustbox}
\newcommand{\methodname}{LifelongMemory}
\usepackage[numbers]{natbib}
\usepackage{xcolor}

\definecolor{blue}{RGB}{79,113,190}
\definecolor{pureblue}{RGB}{0,0,255}

\newif\ifarxiv
\arxivtrue

\let\titleold\title
\renewcommand{\title}[1]{\titleold{#1}\newcommand{\thetitle}{#1}}

\usepackage{abbrv}
\usepackage{graphicx}
\usepackage{booktabs}
\usepackage{hyperref}
\hypersetup{
    colorlinks=true,
    linkcolor=pureblue,
    filecolor=magenta,      
    urlcolor=cyan,
    citecolor=pureblue
}
\usepackage{cleveref}
\begin{document}

\title{\bf LifelongMemory: Leveraging LLMs for Answering Queries in Long-form Egocentric Videos} 
\author{
  Ying Wang, Yanlai Yang, Mengye Ren \\\\
  New York University\\
  \texttt{\{yw3076,yy2694,mengye\}@nyu.edu} \\
  \url{https://agenticlearning.ai/lifelong-memory}
}

\date{}

\maketitle

\begin{abstract}
In this paper we introduce LifelongMemory, a new framework for accessing long-form egocentric videographic memory through natural language question answering and retrieval. LifelongMemory generates concise video activity descriptions of the camera wearer and leverages the zero-shot capabilities of pretrained large language models to perform reasoning over long-form video context. Furthermore, LifelongMemory uses a confidence and explanation module to produce confident, high-quality, and interpretable answers. Our approach achieves state-of-the-art performance on the EgoSchema benchmark for question answering and is highly competitive on the natural language query (NLQ) challenge of Ego4D. Code is available at \url{https://github.com/agentic-learning-ai-lab/lifelong-memory}.
\end{abstract}    
\section{Introduction}

\begin{wrapfigure}{r}{0.5\textwidth}
    \vspace{-0.15in}
    \centering
    \includegraphics[width=0.45\textwidth,trim={0 1.2cm 0 0},clip]{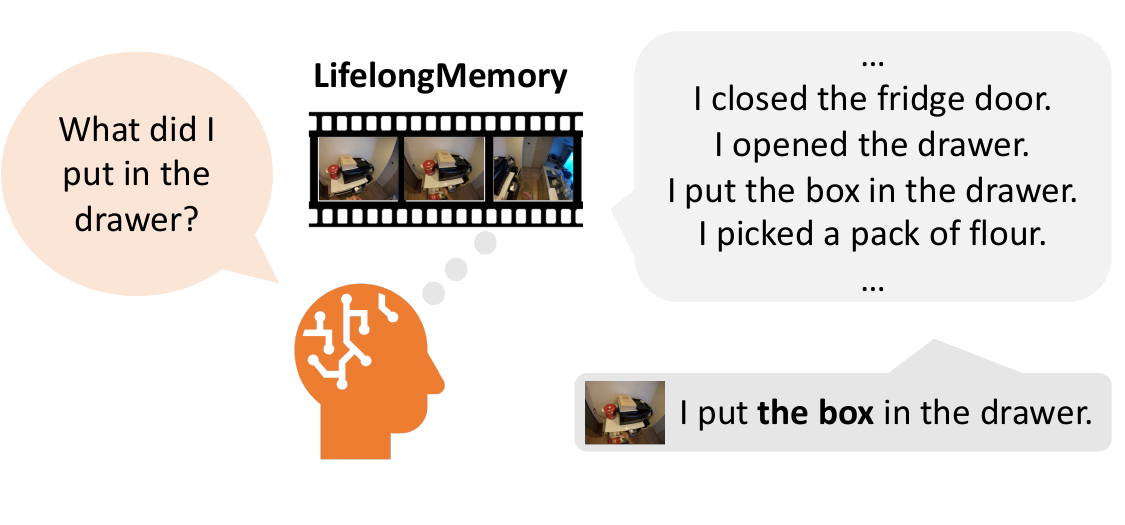}
    \caption{\methodname{} employs natural language descriptions to create an episodic memory. It uses an LLM to sift through past events, retrieving specific moments in response to queries.\vspace{-0.05in}}
\end{wrapfigure}

Long-form egocentric video understanding has the potential to make a tremendous impact in real-life applications such as personalized AI assistants.
Imagine awkward moments when you find yourself asking ``where did I put my glasses'' or ``what is the person's name I just talked to.'' A personalized AI assistant with a video memory can help us search for answers to questions like these. It takes in a question in the form of a natural language, and outputs either an answer or a video playback of the exact moment when the event of interest took place.

However, despite the progress made on video and natural language understanding in deep learning, long-form egocentric video question answering remains challenging for two reasons. First, unlike short-form videos~\cite{kinetics400, Xu_2016_CVPR_MSRVTT, hendricks17iccv_didemo} that usually only contain one single scene and action, long-form egocentric videos can involve multiple scenes where the camera wearers perform numerous tasks and interact with different people and objects. The abundance of details and long-range temporal dependencies make successful information retrieval difficult. Previous methods develop better video features to capture low-level action and object information~\cite{Feichtenhofer_2019_ICCV_slowfast, lin2022egocentric, egovlp, internvideo, shao2023action}, yet fall short of long-form video understanding~\cite{Mangalam2023EgoSchemaAD, ego4d}.
Second, question answering may require sophisticated reasoning of events and oftentimes end-to-end models do not have enough supervision data to generalize and correctly understand different types of questions~\cite{naq}.

\begin{figure*}[t]
  \centering
  \includegraphics[width=0.9\textwidth,trim={0 0 0 0},clip]{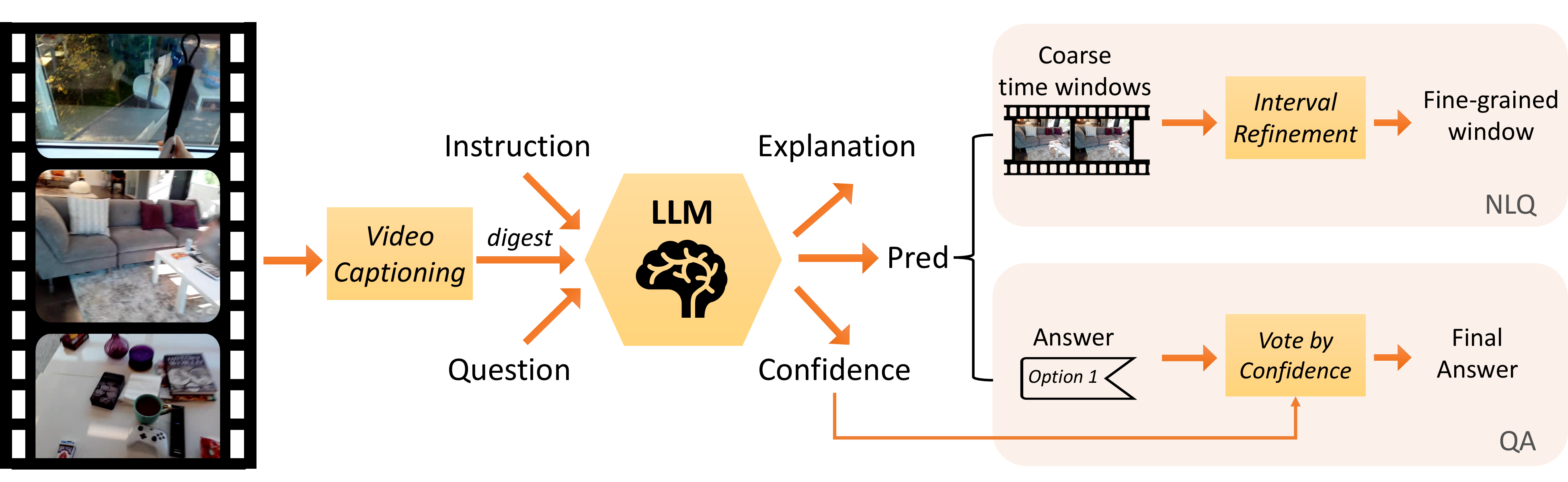}
  \caption{
Our \methodname{} Framework for Long-form Video Understanding. The video inputs are first converted into captions using a pretrained MLLM and then condensed via \emph{Caption Digest}. Next, captions and queries are processed by an LLM to predict coarse temporal windows (for NLQ) or answers (for QA) with explanations and confidence levels. For the NLQ task, the predicted windows are further refined by a pre-trained NLQ model. For the video QA task, we ensemble the predictions of multiple runs and select the answers with the highest confidence. \vspace{-0.05in}}
\label{fig:pipeline}
\end{figure*} 

To address these two challenges simultaneously, we propose a unified framework, \emph{\methodname{}}, for long-form video question answering using large language models (LLMs). We compress long video inputs into concise text descriptions with our proposed \emph{Caption Digest} component. The text format can be then augmented to the context of an LLM for answering the questions and locating the most relevant time window. The LLM is capable of general question answering, and unlike end-to-end models, it has zero-shot generalization. Moreover, we also prompt the LLM to produce a confidence level with a textual explanation, both of which help refine the predictions and enhance the interpretability of the model outputs.

Our proposed framework achieves superior performance on two benchmarks for long-form egocentric video understanding, including multi-choice Video Question Answering (Video QA) and natural language query (NLQ). For zero-shot evaluation on the EgoSchema video QA benchmark~\cite{Mangalam2023EgoSchemaAD}, our method achieves the state of the art which doubles the accuracy of pretrained video QA models~\cite{internvideo,ye2023mplugowl, fu2023empirical_VIOLET, yang2022zeroshot_FrozenBiLM} and significantly outperforms other LLM-based methods~\cite{zhang2023simple_LLoVi, wang2023vamos}. In the Ego4D NLQ challenge, our method is able to increase the precision of pretrained NLQ models~\cite{naq, hou2023groundnlq} by providing coarse-grained candidate temporal windows in zero shot. In summary, our contributions are as follows:
\begin{itemize}
    \item We propose a novel framework, \emph{\methodname{}}, that integrates pre-trained MLLMs to answer questions in long-form egocentric videos. It leverages the remarkable reasoning capabilities of LLMs to tackle the challenge of long-range temporal understanding.
    \item Our method significantly outperforms prior models and concurrent LLM-based solutions on EgoSchema, and remain highly competitive on Ego4D NLQ.
    \item Our framework enhances the interpretability and reliability of the results by providing a confidence level and textual explanation of its prediction, revealing the reasoning process of LLMs.
    \vspace{-0.05in}
\end{itemize}

\section{Related Work}
\label{sec:related_work}

Multimodal Large Language Models (MLLMs) have recently demonstrated their impressive capabilities in various downstream vision-language tasks~\cite{li2023seed, li2023seed2, MMBench}. In this paper, we discuss how to utilize frozen MLLMs for long-form video understanding by experimenting with two specific tasks---Video Question Answering~(Video QA) and Natural Language Queries~(NLQ)---both of which require comprehensive understanding and reasoning of texts and videos. In the following paragraphs, we survey prior works on MLLMs, Video QA, and NLQ.

\vspace{-0.1in}

\paragraph{Multimodal Large Language Models.} Large language models (LLMs)~\cite{gpt3, zhang2022opt, gpt4, llama, llama2, llama3, palm, vicuna} have demonstrated their excellent ability to understand and reason with natural language inputs~\cite{sparks, huang-chang-2023-llm-survey, zhang2024llmmastermindsurveystrategic}. To extend this understanding and reasoning ability beyond text, many prior works have explored incorporating other modalities, especially visual perception, into LLMs. This leads to the rise of multimodal large language models (MLLMs)\cite{wang2024exploringreasoningabilitiesmultimodal,yin2024surveymultimodallargelanguage, zhang2024mmllmsrecentadvancesmultimodal}. LLaVA~\cite{liu2023llava, llava1_5} connects the CLIP visual encoder~\cite{clip} with the language decoder of an LLM and finetune them end-to-end on multimodal instruction-following data, achieving competitive performance in general-purpose visual and language understanding. LaViLa~\cite{lavila} adds visual conditioning on the input to pre-trained LLMs, and finetunes them on Ego4D narrations \cite{ego4d} to create automatic video narrators. These MLLMs serve as critical components in a broad range of downstream applications, showcasing their strength in reasoning on multimodal data~\cite{yang2023mm, zhao2023chat, du2023video, chen2023large, palmE, yang2023mm}. Our work shares the successes in utilizing MLLMs for understanding and reasoning on text and image/video data. Although most current open-sourced MLLMs (such as LLaVA and  LaViLa) only take one image or a very short video as inputs, our proposed framework can integrate those pretrained MLLMs and leverage them for the challenging task of long-form video understanding. 

\vspace{-0.1in}

\paragraph{Video Question Answering with Multimodal LLMs.} The success of LLMs in text QA~\cite{vicuna, llama, llama2, gpt4} leads to an increasing trend of applying MLLMs in video QA tasks~\cite{yang2021just, yang2022zeroshot_FrozenBiLM, zeng2022socraticmodels, wang2022language, pan2023retrieving, wang2023vamos, zhang2023simple_LLoVi, zhang2023multimodal}. Due to the computational burden of large-scale pertaining, many prior works have explored leveraging pretrained (M)LLMs for zero-shot or few-shot QA. R2A~\cite{pan2023retrieving} retrieves textual descriptions from an external text corpus based on the similarity of video frames and text features encoded by CLIP~\cite{clip}, then uses a pretrained LLM to generate answers given the question and the retrieved descriptions. However, retrieval from a pre-defined text corpus hinders scaling to unseen videos and results in vague or inaccurate video descriptions that can decrease the accuracy of the subsequent QA step. Instead of obtaining captions by retrieval, some prior works utilize a pretrained captioning model to generate high-quality descriptions of videos. VidIL~\cite{wang2022language} obtains frame-level captions from BLIP~\cite{blip} and retrieves labels of objects, attributes, and events from pre-defined vocabularies using CLIP~\cite{clip}, then feed all these information into an LLM with a few labeled examples. Despite its good performance on short videos~\cite{xu2017video_msrvttqa_msvdqa}, this approach is not suitable for long videos in the wild because (i) pre-defined vocabularies limit the application of the approach and (ii) a large number of noisy and redundant low-level details can distract the LLM from the main task. Our proposed framework is more efficient and flexible: It is able to perform zero-shot video QA that only utilizes a concise list of distilled captions without requiring a fixed keyword vocabulary. Most relevant to our work, Socratic models~\cite{zeng2022socraticmodels} show qualitative examples of LLM's zero-shot performance on some toy examples, where the key moments of the input video are converted into a textual record by a captioning model. Concurrent work Vamos~\cite{wang2023vamos} and LLoVi~\cite{zhang2023simple_LLoVi} employ a similar approach of using a captioning model to bridge videos and LLMs for zero-shot video QA. Empirical evaluations show that our proposed framework significantly outperforms these two on EgoSchema~\cite{Mangalam2023EgoSchemaAD}, a zero-shot video QA benchmark designed for long-range temporal understanding.

\vspace{-0.1in}

\paragraph{Natural Language Queries in Egocentric Videos.} The Natural Language Queries (NLQ) task involves localizing the temporal window corresponding to the answer to a question in a long video clip. This task is challenging for end-to-end supervised video localization models~\cite{vslnet, 2DTAN_2020_AAAI} due to the sparsity of annotations and the length of videos in the dataset. Prior works have focused on constructing a hierarchical structure, augmenting the NLQ dataset and developing better video features through large-scale pretraining. ReLER~\cite{reler} proposes a novel multi-scale cross-modal transformer architecture, a video frame-level contrastive loss, and two data augmentation strategies. InternVideo~\cite{internvideo} improves the quality of video features by carefully pre-training and fine-tuning a VideoMAE-L Model~\cite{videomae}, and ensemble the features and predictions. More recently, NaQ~\cite{naq} introduces a data augmentation strategy to transform video narrations into training data for the NLQ task, alleviating the problem of sparse annotation. NaQ++ ReLER, obtained by training the ReLER model with NaQ data, was the previous state-of-the-art method for Ego4D NLQ. GroundNLQ~\cite{hou2023groundnlq} is the current state-of-the-art for this benchmark. It adopts a two-stage pre-training strategy to respectively train a video feature extractor and a grounding model on video narrations, and finally finetune the grounding model on annotated data. Our work is complementary to these prior works in that they can be used in the last stage of our proposed framework to produce more fine-grained predictions based on the predictions of the frozen LLM.

\vspace{-0.1in}

\section{LifelongMemory}
In this section, we introduce our proposed \methodname{} framework. To tackle the challenge of long-form videos, we first transform egocentric videos into a comprehensive yet concise textual log and then further condense the information via \emph{Caption Digest}. Then, we use an LLM to predict answers (for Video QA) or coarse temporal windows (for NLQ), along with confidence and explanation for interpretability. Finally, the predictions are further refined depending on the task. \Cref{fig:pipeline} outlines the workflow and we describe different stages in detail below.

\subsection{Egocentric Video Captioning}
\looseness=-10000
We begin by summarizing the raw footage into a list of captions using pre-trained MLLMs (\eg LaViLa \cite{lavila}). 
We sample image frames or short video clips at a fixed interval, and produce a line of caption per clip.
The text descriptions as a form of episodic memory enable the transformation of complex egocentric video footage into a coherent log of daily activities, capturing life's narrative in a more accessible and compressed format.

\begin{figure*}[t]
  \centering
  \includegraphics[width=\linewidth]{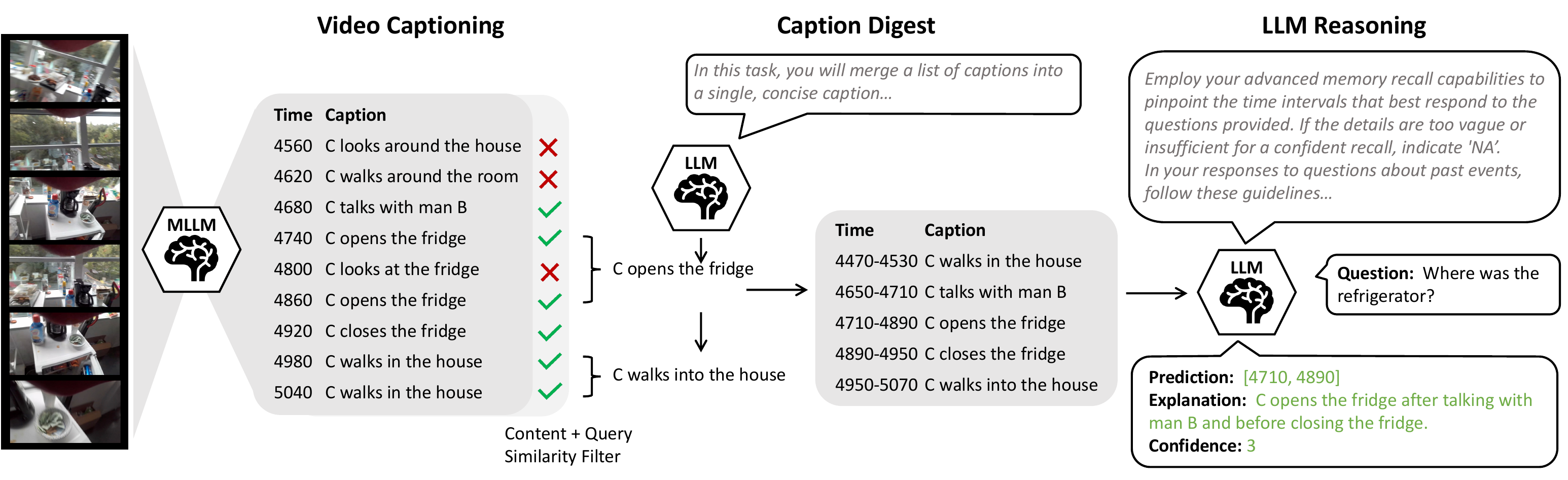}
  \caption{
  Example Caption Processing and LLM Reasoning for NLQ. 1) We use a multimodal LLM (MLLM) to produce captions from a list of short video clips.
  2) Content and query similarity filters are then applied to remove redundant and irrelevant captions. Similar consecutive captions are merged by an LLM. 3) An LLM is instructed to take inputs from the list of condensed captions and retrieve the most relevant interval candidates. The same procedure is performed on the QA task.
  }
  \label{fig:detail}
\end{figure*}

\paragraph{Caption Digest.}
\looseness=-10000
Raw captions produced by MLLMs, however, can be rather verbose and repetitive, and consequently hinder the downstream reasoning process, especially for long-form videos. We propose to create a caption digest to condense the information. Moreover, we aim to increase the relevance of the captions in relation to the target queries. \Cref{fig:detail} shows an example of the caption digest process.
First, we remove uninformative captions (\eg ``looks around ...''). Second, we remove captions that are not relevant to the query by comparing the embedding similarity. Third, we gather adjacent captions that share a high similarity score and use an LLM to produce a single concise caption. The condensed list of captions then augments the context of the LLM for further reasoning and processing.

\subsection{LLM Reasoning}
With the list of condensed captions with their corresponding time interval from the previous stage, we leverage an LLM here for its impressive zero-shot context understanding and reasoning capability.
We combine captions and queries into an instructive and contextualized prompt. A snippet of the instruction template is shown in \Cref{fig:detail}. The full prompt and a discussion of the prompt designs are in~\Cref{prompt}.

We particularly instruct the LLM to aggregate information and imagine the visual scene underlying the given captions. The LLM is instructed to take into consideration the full context in the template and utilize different pieces of information to produce the most probable answer.
For example, when asking ``\emph{Who did I interact with when I was shopping?}'', the LLM is able to filter all captions and produce a list of intervals involving ``\emph{person x talking to C}'' where \emph{C} is the subject in the video and \emph{X} refers to the other person. The LLM is also instructed to consider the loss of information when converting videos into concise captions. For example, one query asks ``\emph{What size of washer did I pick ?}'' but there are no captions explicitly mentioning the washers. In this example, the LLM displays its capability to capture implicit information and infer based on context. The LLM answers ``\emph{choosing the time points where I picked items from the table or the floor, as these instances may provide more context about the objects and their locations}.'' By grasping nuanced relationships and dependencies within the given context, LLM is able to filter out the most relevant information from the extensive video captions.

In addition to the predicted answers, we also ask the LLM to explain its predictions for more interpretability. Specifically, we ask the LLM to output a sentence of explanation to encourage introspective thinking and rate its confidence in the output out of three confidence levels. The verbalized confidence strategy~\cite{xiong2023can} can help us control the precision of the output in later stages.

\subsection{Vote by Confidence (Video QA)}
\label{sec:votebyconf}

To increase the reliability of LLM predictions for video QA, we ensemble
the LLM's predictions using voting by confidence. We repeatly perform the LLM reasoning step where the LLM is prompted to generate predictions based on the same input in each run. From the pool of predictions, the answer with the highest confidence score is selected. In cases where multiple answers have the same highest confidence, a random selection is performed. By focusing on the most confident predictions, this ensemble step can further improve the accuracy and robustness of the results.

\subsection{Fine-grained Interval Refinement (NLQ)}

Since the time intervals are subsampled, to obtain a fine-grained interval prediction for the NLQ task, we revisit the video inputs and enhance our LLM interval predictions in the last stage.
For this goal, we employ a pretrained NLQ model and feed in candidate intervals predicted by our previous stage. The intervals are padded with a small window of size $\alpha$.
Specifically, for each $(s_i,e_i)$, the new start time is $s'_i = \max(s_i-\alpha, s)$ and the new end time is $e'_i = \min(e_i+\alpha, e)$ where $s$ and $e$ are the start and end time of the original clip. Then we extract video clips $[v_1,v_2 \dots v_n]$ according to the predicted intervals $[(s'_1,e'_1),(s'_2,e'_2) \dots (s'_n,e'_n)]$. 

\looseness=-10000
When the prediction for a certain query contains multiple candidate intervals, we feed them along with the target query into a classifier that is trained on NLQ data to select the optimal candidate $v^*$. For queries without predictions (\ie ``NA''), we simply use the original full video. Localization within a coarse temporal window makes the NLQ task easier compared to doing it on the original full-length video.

\section{Experiments}

In this section, we evaluate our LifelongMemory framework in real-world egocentric video query tasks.

\subsection{Experiment Setup}
\label{experiment setup}

\paragraph{EgoSchema.} The EgoSchema dataset \cite{Mangalam2023EgoSchemaAD} consists of over 5,000 question-answer pairs for 250 hours of Ego4D videos covering a wide range of human daily activities. For each question, the correct answer needs to be selected from 5 choices based on a three-minute egocentric video. The dataset is curated by human annotators to ensure all questions require long-term temporal understanding. We use the subset provided by EgoSchema, which contains 500 question-answer pairs, for ablation studies on prompt designs, captioning choices, and voting by confidence, then use the best setup for evaluation on the full benchmark.

\paragraph{Ego4D NLQ.} The Ego4D dataset \cite{ego4d} is an egocentric video dataset including a wide variety of daily life activities recorded by individuals wearing cameras. The NLQ task, as one of the episodic memory tasks of Ego4D, requires localizing a temporal window of the video to answer a natural language query. The NLQ annotations are from 227 hours of videos, with a total of 19,200 queries spanning 13 query templates. The train/val/test split (60\%, 20\%, 20\%) is composed of disjoint sets of video clips. The average video length is approximately 8.7 minutes, while the average duration of a response window is only 9.3 seconds, representing on average only 2\% of the full video.

\paragraph{Evaluation Metrics.} For the EgoSchema dataset, we use accuracy to evaluate our framework since it is a multi-choice QA task. For the NLQ dataset, we adopt different metrics for different stages as below. In the LLM Reasoning stage where we only have coarse-grained predictions, we evaluate on the validation set with metrics including \textbf{(i)} the ratio of predictions that overlap with the ground truth (denoted as \emph{Overlap}), \textbf{(ii)} and the proportion of predictions where at least one candidate achieves an Intersection over Union (IoU) greater than 0.3 with the ground truth (denoted as \emph{IoU*@0.3}). During the refinement stage for NLQ, we obtain fine-grained predictions so we can evaluate the test dataset using the standard NLQ metrics -- \emph{R@1 IoU@0.3} and \emph{R@1 IoU@0.5}, which is the recall of top one prediction having IoU with the ground truth larger than the threshold \{0.3, 0.5\}.

\paragraph{Caption Sources.} We experiment with machine-generated captions and human-annotated captions and test the effect of text-conditioned captioning.
\begin{itemize}
    \item \textbf{LLaVA}: LLaVa~\cite{llava1_5,liu2023llava} is a multimodal LLM pre-trained on a diverse set of 1.2M publicly available data, including various multimodal question-answering and reasoning tasks. To encourage LLaVA to generate captions that are relevant to the query while not introducing false positives, we follow the template proposed by LLaVA-1.5\cite{llava1_5} and adopt the prompt \emph{``If there are factual errors in the questions, provide a precise description of the image; if not, proceed answering the question. [queries].''}
    \item \textbf{LaViLa}: LaViLa~\cite{lavila} is a multimodal LLM pre-trained on the video-narration pairs from Ego4D and is thus capable of generating captions that mimic the ground-truth descriptions of the video. Each caption is generated using 4 frames uniformly taken from a two-second video clip.
    \item \textbf{Ego4D Narrations}: Ego4D~\cite{ego4d} narrations include written sentence narrations in English from human annotators, describing a diverse set of activities in the dataset. The annotated narrations contain on average 13.2 sentences per minute of video, which is not as dense as the LLaVA and LaViLa captions that we sample every 2 seconds.
    
\end{itemize}

\paragraph{Caption Digest Details.}
The generated captions are distilled through filtering and merging in this step. We first remove ambiguous captions containing keywords that are associated with blurry and noisy frames. Then, we filter out irrelevant captions based on the similarity scores between the embedding of queries and captions encoded by LaViLa. Lastly, we identify groups of similar consecutive captions by calculating the similarity scores of the embeddings of neighboring captions and merge captions in the same group by querying GPT-3.5 with prompt \emph{``In this task, you will merge a list of captions into a single, concise caption. Focus on clarity and brevity while ensuring no critical details are lost in the merging process.''} 

\paragraph{NLQ Refinement Details.}
For the refinement stage, we train a classifier on the NLQ train set to select the optimal candidate from multiple LLM predictions. To construct a video dataset similar to the real LLM predictions, we randomly shift and scale the ground-truth temporal windows. 
We then mark those intervals that have IoU with the ground truth larger than 0.5 as positives and randomly pick the same amount of negative samples from intervals with IoU less than 0.1. We utilize video features encoded by InternVideo~\cite{internvideo} and EgoVLP~\cite{egovlp} and adapt VSLNet~\cite{vslnet}, a span-based localization network, to this video classification task, where we replace the localization head with a classification head. 
After obtaining the optimal candidate temporal windows, we extend them by a window size of $\alpha$ to provide more context to the NLQ model and then feed them into the state-of-the-art NLQ model, NaQ++ReLER \cite{naq, reler} and GroundNLQ \cite{hou2023groundnlq}, which have been finetuned on the Ego4D NLQ dataset. This gives us fine-grained predicted temporal windows that can reflect the answers to the target queries.

\begin{figure*}[h!]
\centering
\includegraphics[width=0.9\textwidth,trim={0 0.75cm 0cm 0cm}]{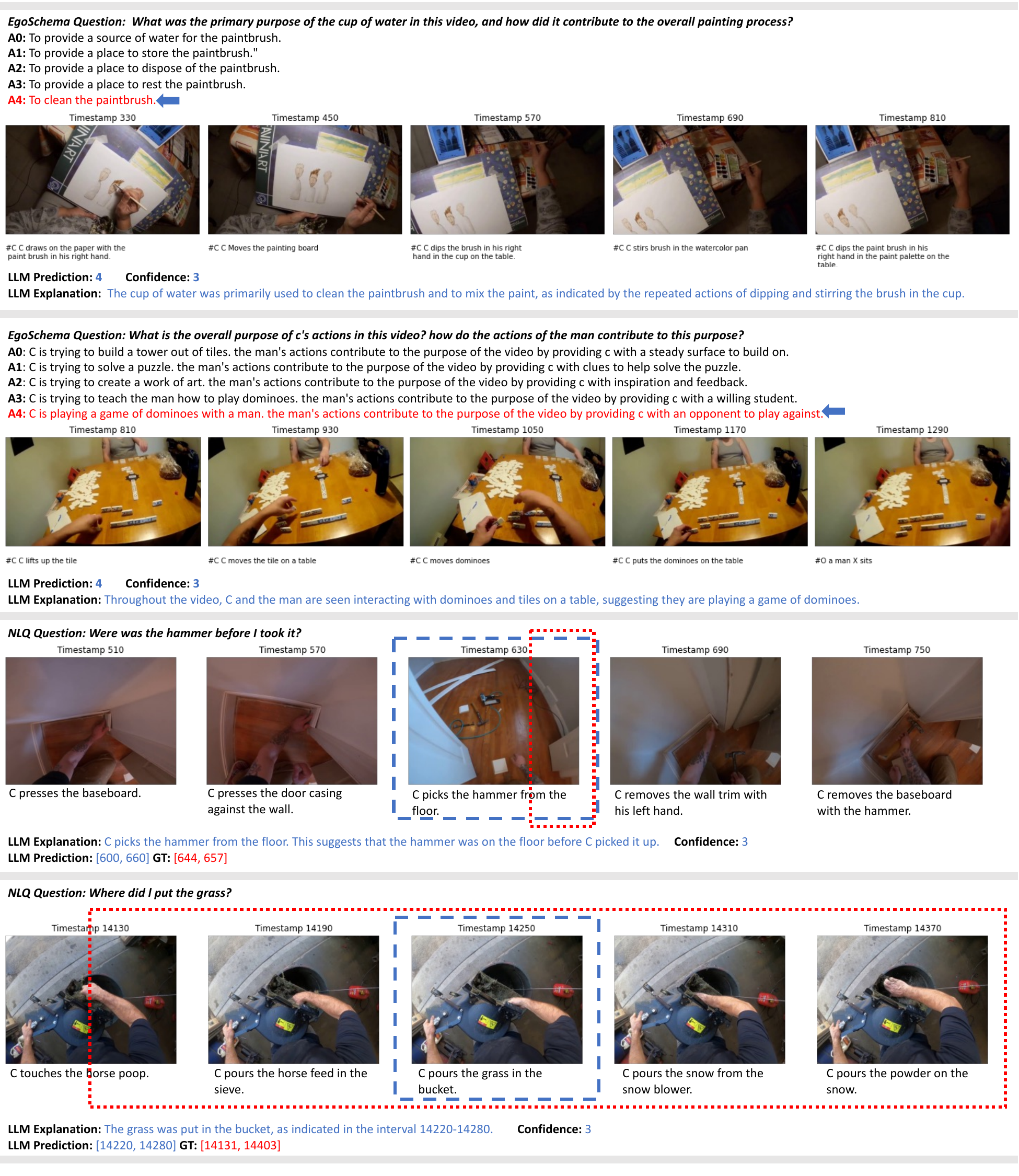}
\caption{EgoSchema QA and Ego4D NLQ examples using LaViLa and GPT-4. The groud-truth answers are in \textcolor{red}{red} and the LLM predictions are in \textcolor{blue}{blue}. The sampled frames are manually picked from the raw video input to show key events related to the query.}
\label{fig:viz_main}
\end{figure*}

\subsection{Qualitative Results}
We visualize \emph{\methodname{}} results in Figure~\ref{fig:viz_main}. For the EgoSchema QA examples, we show that the captions capture objects and actions in the scene very well, and the LLM is then able to answer the question correctly, using its reasoning capability. For the NLQ examples, we show that many LLM predictions have high-quality overlaps with the ground-truth windows (without interval refinement). We note that many successful retrievals rely on high-quality captions and we expect there can be a large room for future improvement with a stronger captioning model. In both datasets, the LLM is able to explain its predictions with a confidence level, enhancing the interpretability of the results. We provide more qualitative examples in \Cref{more_qual}.

\subsection{Quantitative Results}
\label{sec:results}

\paragraph{EgoSchema Benchmark Results.} Our method achieves state-of-the-art performance on EgoSchema, as shown in~\Cref{tbl:egoschema}. Due to the challenge of long-form videos, prior state-of-the-art video QA models~\cite{internvideo, yang2022zeroshot_FrozenBiLM} struggle at this task with an accuracy not much better than random (20\%). When compared with concurrent works -- LLoVi~\cite{zhang2023simple_LLoVi} and Vamos~\cite{wang2023vamos} -- that also leverage GPT-4, our approach outperforms them with a significant margin of over 10\%. These empirical results confirm \methodname{} is a simple yet effective framework that can reason and answer questions of very long egocentric videos.

\begin{table}[t]
\footnotesize
\centering
\caption{Zero-shot QA on EgoSchema with Different Models. *~represents ensembled results using vote by confidence. Subset is 500 question-answer pairs provided by EgoSchema for validation. }
\resizebox{0.7\columnwidth}{!}{
\begin{tabular}{c|cccc}
\toprule
\textbf{Model} & \textbf{\; LLM \; }  & \textbf{\; Input \; } & \textbf{\;  Subset \;} & \textbf{\;  Full \;}  \\
\midrule
FrozenBiLM ~\cite{yang2022zeroshot_FrozenBiLM} & - & 90 frames & - & 26.9 \\
InternVideo~\cite{internvideo} &  - & 90 frames & - & 32.1 \\
Vamos~\cite{wang2023vamos} & GPT-4 &  mixed & 51.2 & 48.3  \\
LLoVi~\cite{zhang2023simple_LLoVi} & GPT-3.5 & 180 captions & 57.6 & 50.3  \\
LLoVi~\cite{zhang2023simple_LLoVi} & GPT-4 & 180 captions & 58.3 & -  \\
Ours & Llama3-8B &  90 captions & 60.4 & - \\
Ours & GPT-3.5 &  90 captions & 64.0 & - \\
Ours & Claude-3-Haiku &  90 captions & 64.8 & 55.2 \\
Ours & GPT-4 &  90 captions & 68.0 & 62.1\\
Ours* & GPT-4 &  90 captions & 69.0 & 62.4 \\
Ours & GPT-4o &  90 captions & 70.6 & 64.6 \\
Ours* & GPT-4o &  90 captions & \textbf{72.0} & \textbf{64.7} \\
\bottomrule
\end{tabular}
}
\label{tbl:egoschema}
\end{table}

\paragraph{Ego4D NLQ Benchmark Results.}
We compare the performance of our method and two other competitive methods on the Ego4D NLQ benchmark\footnote{Our results of GroundNLQ are slightly lower than their reported numbers. Since GroundNLQ has not released all checkpoints, we are unable to reproduce the results.} in~\Cref{tbl:result_nlq}. Our method with Ego4D ground-truth narrations achieves the best performance in the validation set and our method with GroundNLQ as the refinement model achieves the best performance in the test set. \methodname{} is a flexible framework that can be plugged into any pretrained captioning model and video localization model, suggesting the potential of our method for future improvement using better pretrained MLLMs.

\begin{table}[t]
    \centering
    \caption{Ego4D NLQ benchmark results, using GPT-4. Our approach filters out noisy content in the video for the pretrained NLQ models, increasing the precision of the predictions of the NLQ models. Reported metrics all use predictions that rank the first.}
    \resizebox{0.6\textwidth}{!}{
    \begin{tabular}{l|rrrr}
    \toprule
    \textbf{Method} & \textbf{Set} & \textbf{Mean} & \textbf{IoU=0.3} & \textbf{IoU=0.5}\\
    \midrule
    NaQ++ \cite{naq} & val & 20.20 & 25.00 & 15.40 \\
    Ours (LaViLa, NaQ++) & val & 19.00 &	23.40 &	14.61   \\
    Ours (Ego4D, NaQ++) & val & \textbf{21.09} & \textbf{26.12} & \textbf{16.06} \\
    \midrule
    NaQ++ \cite{naq} & test & 17.67 & 21.70 & 13.64 \\
    Ours (LaViLa, NaQ++) & test & 18.06 & 22.28 & 13.84 \\
    GroundNLQ \cite{hou2023groundnlq} & test & 20.08 & 23.43 & 16.71 \\
    Ours (LaViLa, GroundNLQ) & test & \textbf{20.27} & \textbf{23.68} & \textbf{16.86} \\
    \bottomrule
    \end{tabular}
    }
    \label{tbl:result_nlq}
\end{table}

\begin{table*}[t]
\centering
\caption{NLQ performance using different caption and LLM components. The \textbf{bold} number denotes the highest and the \underline{underlined} the second highest. $\dagger$ represents predictions with a confidence level of 3. \emph{NA} represents the ratio of null predictions, and all other metrics do not include null predictions. \emph{Count} represents the number of captions, and \emph{Length} represents the average word count of captions.}
\resizebox{0.9\textwidth}{!}{
\begin{tabular}{l|cccrrrrrr}
\toprule
\textbf{Captions} & \textbf{\emph{Count}} & \textbf{\emph{Length}} & \textbf{LLM} & \textbf{\emph{Overlap}} & \textbf{\emph{Overlap}} $\dagger$ & \textbf{\emph{IoU*@0.3}} & \textbf{\emph{IoU*@0.3}} $\dagger$ & \textbf{\emph{NA}}  \\
\midrule
\multirow{2}{*}{Ego4D} & \multirow{2}{*}{109} & \multirow{2}{*}{7.95} & GPT-4 & \textbf{51.73} & \textbf{53.98} & \textbf{15.99} & \textbf{27.38} & 40.29 \\
 &  & & GPT-3.5 & 31.27 & 33.15 & 0.91 & 4.34 & 94.13 \\
\midrule
\multirow{2}{*}{LaViLa} & \multirow{2}{*}{186} & \multirow{2}{*}{6.40} & GPT-4 & \underline{36.61} & \underline{38.35} & \underline{9.74} & \underline{19.22} & 47.04 \\
 &  &  & GPT-3.5 & 20.47 & 22.33 & 1.29 &  4.78 & 89.75 \\
\midrule
\multirow{1}{*}{LLaVA} & \multirow{1}{*}{250} & \multirow{1}{*}{52.52} & GPT-4 & 6.42 &  8.79  &  1.50  & 2.71  & 60.92  \\
\bottomrule
\end{tabular}
}
\label{tbl:model_comp}
\end{table*}

\begin{table*}[t]
\caption{Ego4D NLQ and EgoSchema QA performance using LaViLa + GPT-4, with different frame sampling intervals and digest strategy. }
\centering
\resizebox{0.9\textwidth}{!}{
\begin{tabular}{cc|ccccc|cc}
\toprule
 &  & \multicolumn{5}{c}{\textbf{Ego4D NLQ}} & \multicolumn{2}{|c}{\textbf{EgoSchema}} \\
\textbf{Freq.} & \textbf{Digest} & \textbf{\# Captions} & \textbf{\emph{Overlap}} & \textbf{\emph{Overlap}$\dagger$} & \textbf{\emph{IoU*@0.3}} & \textbf{\emph{IoU*@0.3}$\dagger$} & \textbf{\# Captions} & \textbf{\emph{Acc}} \\
\midrule
4s & Yes & 70 & 33.55 & 36.99 & 6.46 & 14.64 & 39 & 26.4  \\
2s & Yes & 186 &\textbf{36.61} & \textbf{38.35} & \textbf{9.74} & \textbf{19.22} & 75 & \textbf{68.0}\\
2s & No & 250 &23.71	& 23.89 & 4.92 & 11.28 & 90 & \textbf{68.0} \\
\bottomrule
\end{tabular}
}
\label{tbl:result_cap}
\end{table*}

\paragraph{Captioning Model Choices.}
We compare the effect of different caption sources in~\Cref{tbl:model_comp}. Although LLaVA generates longer captions conditioned on the queries, the performance of LaViLa is significantly better than LLaVA. This indicates the necessity of adopting an egocentric captioning model that focuses on the core activity of the individual. Despite the effectiveness of LaViLa in this task, we identify that LaViLa tends to generate false positive captions as it is finetuned on Ego4D data. We thus evaluate the ground-truth captions provided by the Ego4D Narrations data and observe that it achieves the best performance with significantly fewer captions. This confirms our assumption that an accurate well-crafted set of captions can effectively summarize the information of the camera wearer's activity in egocentric videos.

\paragraph{LLM Choices.} We compare the effect of different LLMs for EgoSchema and NLQ respectively in \Cref{tbl:egoschema} and \Cref{tbl:model_comp}. We observe that GPT-4 and GPT-4o significantly outperform GPT-3.5 and open-source models like Llama~\cite{llama, llama2, llama3} and Vicuna~\cite{vicuna} for both datasets. Note that the performance drop caused by weaker LLMs is much larger for the NLQ task because this task requires more precise instruction following capabilities: weaker models often misunderstand the prompt and output an answer instead of a list of temporal intervals, leading to a high NA ratio. As our framework is agnostic to LLMs---it's very easy to plug in a future version of LLMs to further boost the performance.

\paragraph{Caption Digest.} We evaluate the effect of caption digest in~\Cref{tbl:result_cap}. With \emph{Caption Digest}, we discard ambiguous and irrelevant captions and use LLM to merge similar ones as described in~\Cref{experiment setup}. For NLQ, this technique significantly improves both metrics by around 10\%, suggesting that a concise context leads to a much better retrieval performance. However, similar effects are not observed in EgoSchema as the original undigested context length is already relatively small (\ie less than 100 captions). Since reduced context lengths save the computation costs, we adopt caption digest for both datasets.

\begin{wraptable}{R}{0.5\textwidth}
    \caption{Effect of explanation and confidence levels.}
    \begin{tabular}{c|rr|r}
    \toprule
    &  \multicolumn{2}{c}{\textbf{Ego4D NLQ}} & \multicolumn{1}{|c}{\textbf{EgoSchema}} \\
    \textbf{Conf. Level} & \textbf{\emph{Overlap}} & \textbf{\emph{IoU*@0.3}} & \textbf{Acc.} \\
    \midrule
    $\geq$ 1 &  36.46 & 9.63 & 68.0\\
    $\geq$ 2 &  36.49 & 17.52 & 69.7 \\
    $\geq$ 3 &  \textbf{38.20}  & \textbf{19.06} & \textbf{74.6} \\
    \midrule
    \textbf{Explanation} & \textbf{\emph{Overlap}} & \textbf{\emph{IoU*@0.3}} & \textbf{Acc.}  \\
    \midrule
    No &  32.73 &	8.65 & 64.2 \\
    Yes & \textbf{36.61}	& \textbf{9.74} &  \textbf{68.0} \\
    \bottomrule
    \end{tabular}
    \label{tbl:result_confidence}
\end{wraptable}

\paragraph{Caption Sampling Interval.} Given the same captioning models and preprocessing process, smaller caption intervals lead to higher performance as they provide richer contexts for the LLM. Since each Ego4D video contains a large amount of activities, coarse-grained captioning is very likely to miss key moments and results in a loss of information. Decreasing the sampling frequency of captioning leads to a large drop in the accuracy of predictions of both NLQ and EgoSchema, as shown in~\Cref{tbl:result_cap}. It is worth noting that using very limited captions leads to a very low EgoSchema accuracy that is not much better than random guess (20\%) due to the significant information loss.

\paragraph{Effect of Explanation.} We also experiment with different prompts in \Cref{tbl:result_confidence}. To encourage LLM reasoning step by step, we provide detailed instructions on how to retrieve the temporal windows and answer the queries while explicitly asking it to explain its prediction. The request for explanation encourages the LLM to reason step by step and improves the performance in both datasets. Moreover, providing textual explanations also increases the interpretability and reliability of the model outputs.

\paragraph{Effect of Confidence Levels.} To encourage the LLM to make more reliable predictions, we also explicitly ask the LLM to predict a confidence level for each of its own outputs. We report the relationship between scores and their confidence levels in \Cref{tbl:result_confidence}. The increase in confidence scores leads to an increase in accuracy in both datasets, suggesting the verbalized confidence scores are calibrated. For EgoSchema, we also use confidence level to vote during model ensembling, leading to a 0.1-0.3\% increase in test accuracy as shown in~\Cref{tbl:egoschema}. 

\subsection{Error Analysis}
\begin{figure*}[ht]
\centering
\includegraphics[width=0.95\textwidth,trim={0 0 0 0},clip]{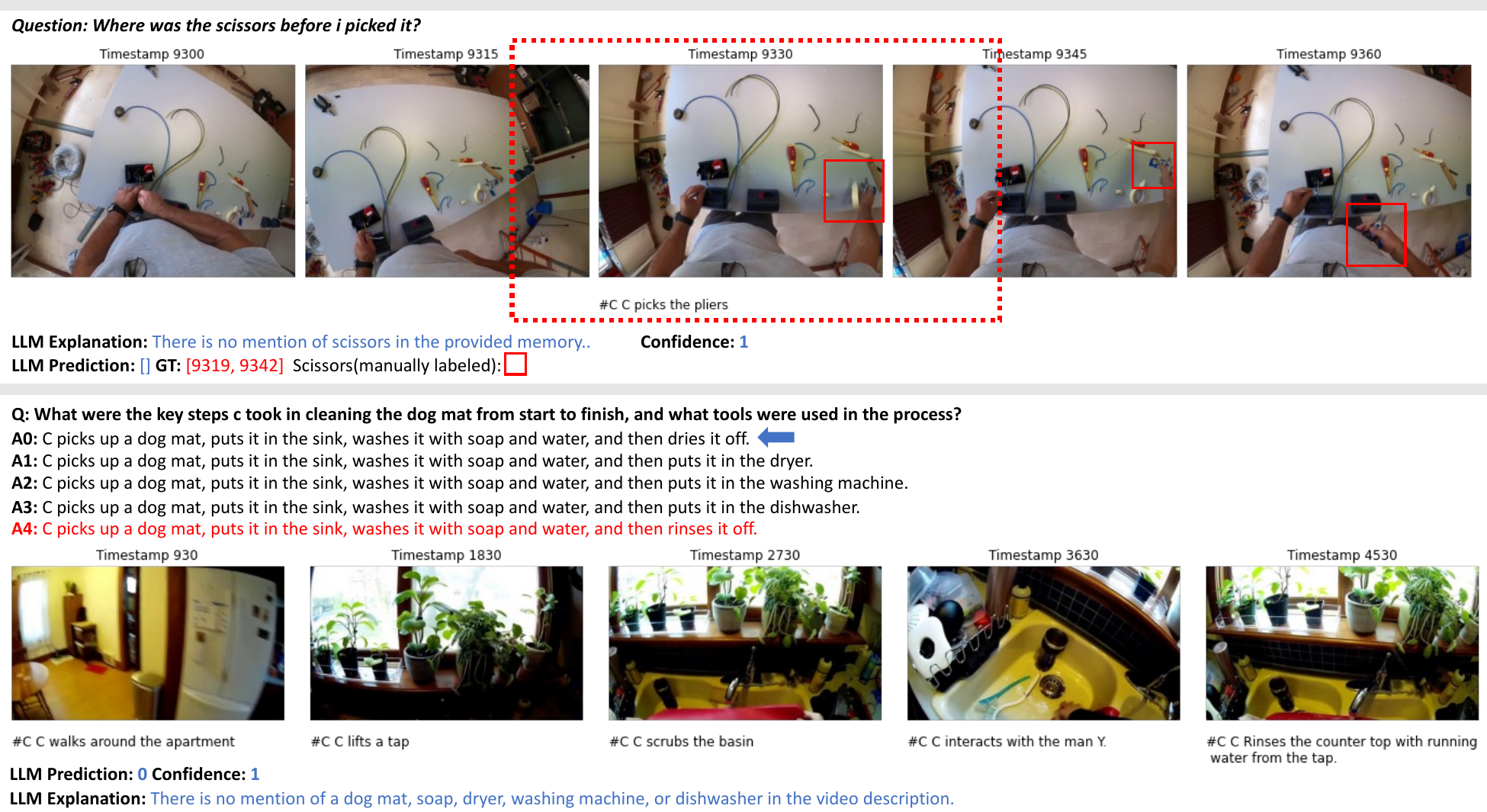}
\caption{Error caused by insufficient captioning. The upper figure is an NLQ example and the lower figure is an EgoSchema example. LLM predictions are in \textcolor{blue}{blue} boxes and the ground truth is in \textcolor{red}{red}.\vspace{-0.1in}}
\label{fig:error}
\end{figure*}

The majority of errors stem from the captioning step, where inevitable information loss occurs during the transformation from long video inputs into text, as shown in \ref{fig:error}. For NLQ with insufficient information, we encourage the LLM to make null predictions and rely on the refinement stage to make the final prediction based on the full input video. On the contrary, we encourage the LLM to select the most plausible answer for EgoSchema when uncertain because we don't rely on a pretrained QA model in the refinement stage. Our prompts are included in~\Cref{prompt}.

We also observe that sometimes the LLM proposes multiple temporal windows for NLQ that seem to be reasonable, but only one ground-truth answer is available, as shown in~\Cref{ambiguity_in_nlq}. This suggests some NLQ queries are ambiguous and require more careful annotations.

\section{Conclusion}
\hyphenation{Ego-Schema}
In this paper, we propose \methodname{}, a novel framework that leverages pre-trained MLLMs for answering natural language queries in long-form egocentric videos. To address the challenges of long-range temporal dynamics, we condense the input videos into a concise textual log and utilize an LLM to comprehend the context and answer the given queries. Our method achieves state-of-the-art performance on EgoSchema and remains highly competitive on Ego4D NLQ, with enhanced interpretability provided by verbalized confidence and explanation. \methodname{} showcases the potential of leveraging LLMs in video understanding and opens up opportunities for personalized AIs that can answer daily queries for individuals requiring assistance.
\section*{Acknowledgment}
We would like to thank the Microsoft Accelerating Foundation Models Research program for providing cloud compute credits for running some parts of our LLM experiments. This work was also supported in part through the NYU IT High Performance Computing resources, services, and staff expertise.

\bibliographystyle{bibstyle}
\bibliography{main}

\clearpage
\appendix 

\section{Prompting}
\label{prompt}
\begin{figure}[ht]
\centering
\includegraphics[width=\textwidth]{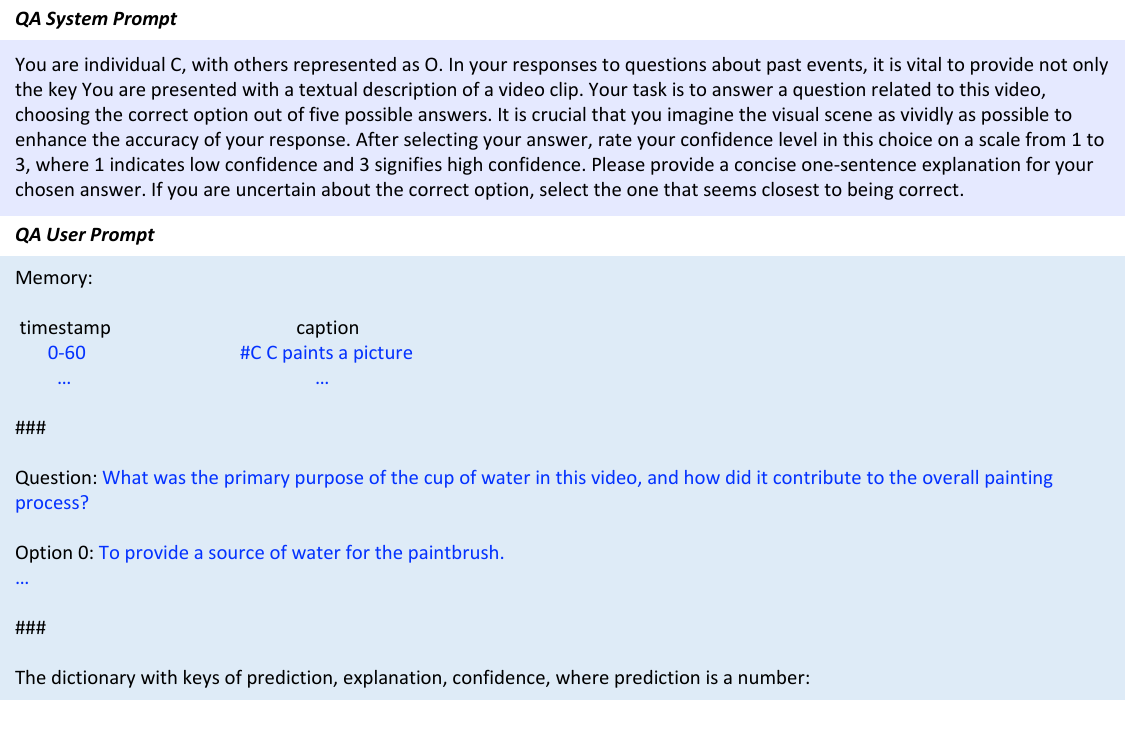}
\vspace{-0.4in}
\caption{System prompt and user prompt for video QA (EgoSchema). The text in \textcolor{pureblue}{blue} should be replaced by the captions and the corresponding question-answer pair.}
\label{fig:prompt_qa}
\end{figure}

\begin{figure}[ht]
\centering
\includegraphics[width=\textwidth]{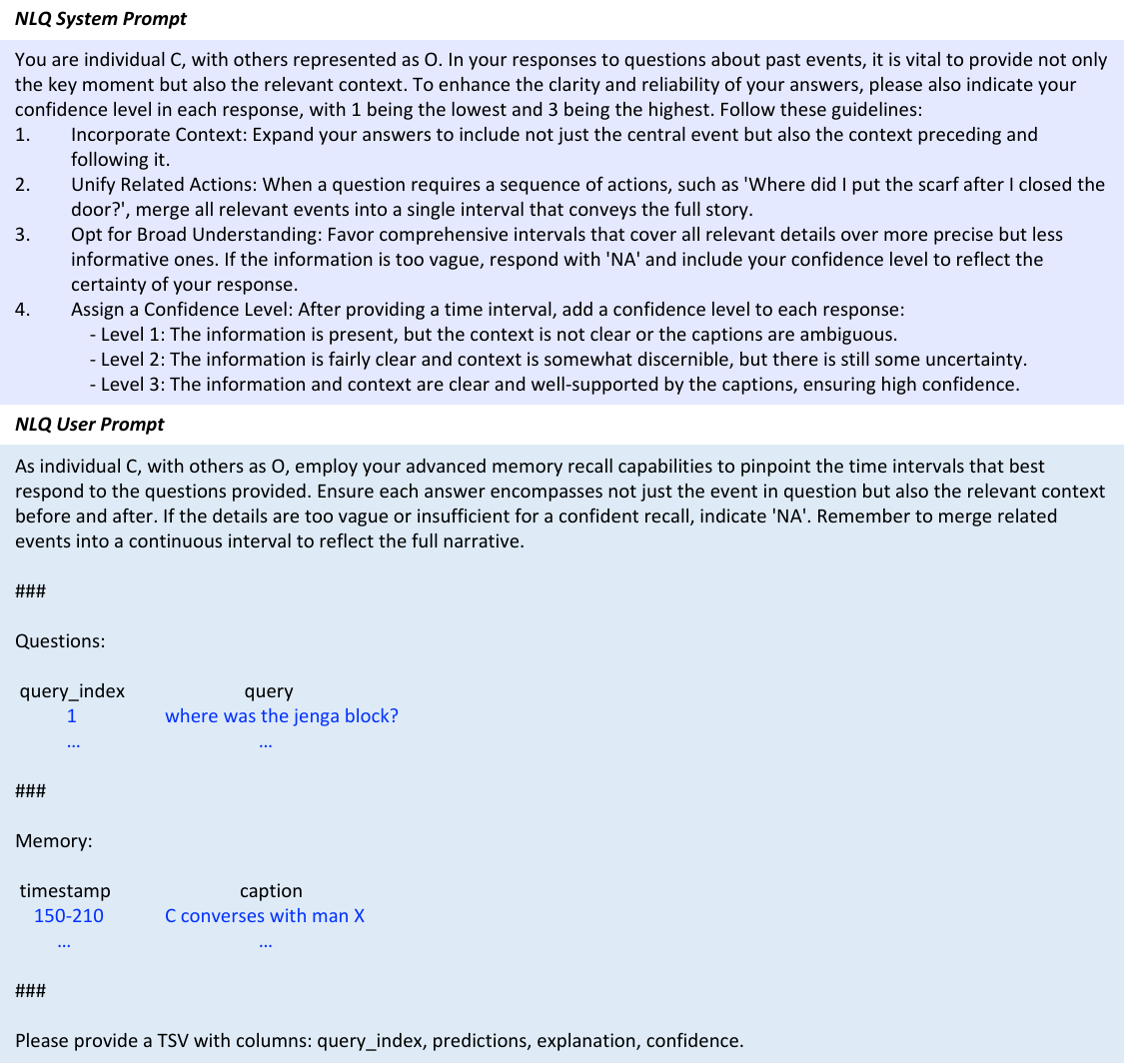}
\caption{System prompt and user prompt for NLQ. The text in \textcolor{pureblue}{blue} should be replaced by the queries and captions in the video clip.}
\label{fig:prompt_nlq}
\end{figure}

We provide the complete prompt used for EgoSchema (\Cref{fig:prompt_qa}) and Ego4D NLQ (\Cref{fig:prompt_nlq}). As the NLQ dataset contains multiple queries for one video clip, we avoid passing the same caption list multiple times by including all queries of the same clip in the prompt to reduce the cost of API calls. We provide an instructive prompt with detailed steps and ask the LLM to produce responses in the structured format to expedite post-processing. Note that we encourage the LLM to refuse to answer NLQ questions if the context is not informative so we can feed the full-length video into the refinement stage later. In contrast, we encourage the LLM to pick the most possible answer for EgoSchema because we must provide an answer to each question and there is no refinement stage for the QA task.

\clearpage

\clearpage

\section{Additional Qualitative Results}
\label{more_qual}
\begin{figure}[h]
\vspace{-0.1in}
\includegraphics[width=\textwidth,trim={0 0 0 0},clip]{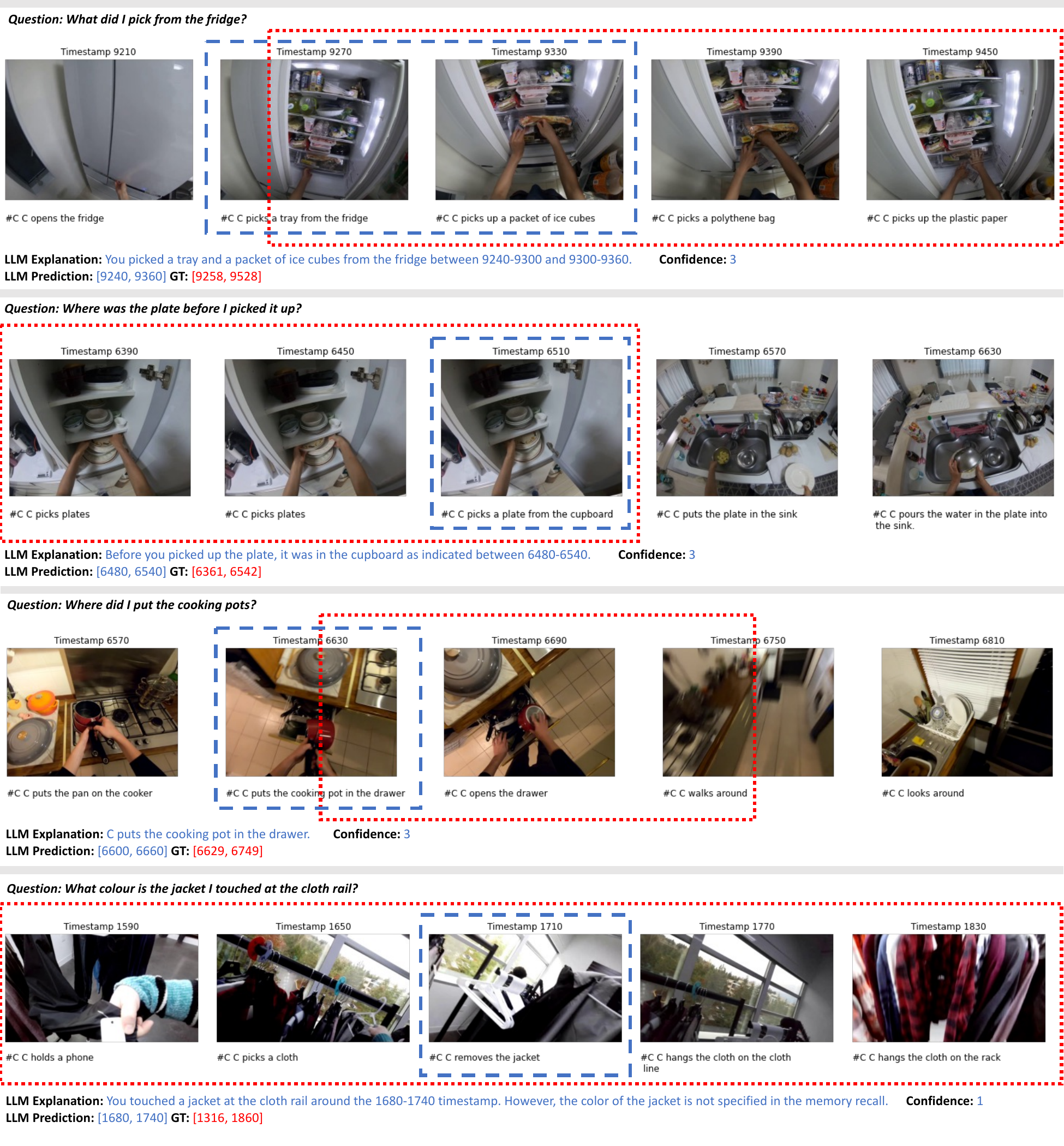}
\vspace{-0.3in}
\caption{NLQ Examples. Each figure represents a two-second 30fps video clip (which is 60 frames). LLM interval predictions are denoted as \textcolor{blue}{blue} boxes and the ground truth is in \textcolor{red}{red}. LLM engine here is GPT-4 and the captioning model is LaViLa. To illustrate the reasoning skills of LLMs, we show the raw LLM predictions without any refinement.}
\label{fig:viz_nlq2}
\vspace{-0.6in}
\end{figure}

\begin{figure}[h]
\vspace{-0.2in}
\centering
\includegraphics[width=\textwidth,trim={0 3cm 0 0},clip]{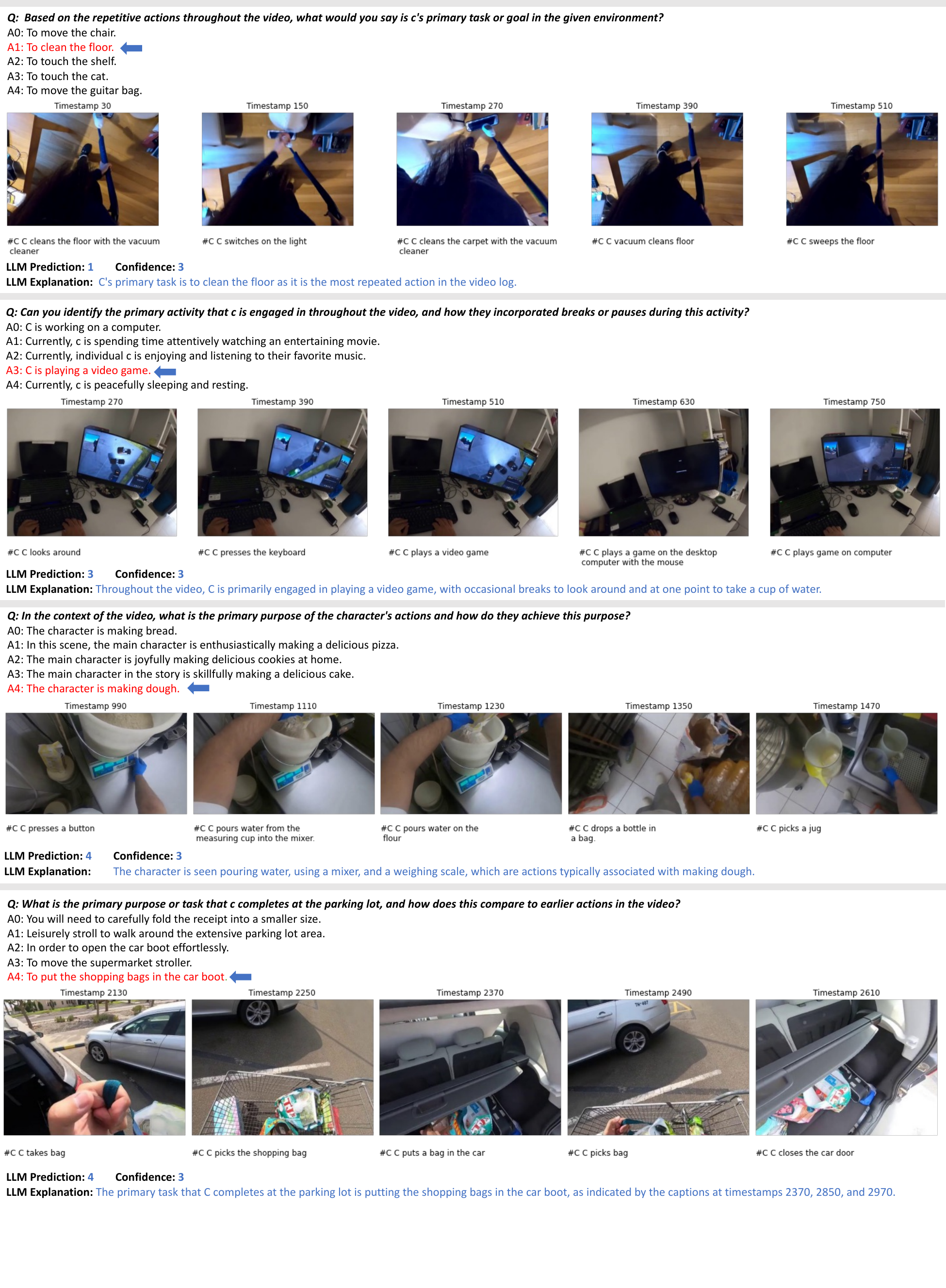}
\caption{EgoSchema Examples. Each figure represents a two-second 30fps video clip (which is 60 frames). LLM predictions are denoted as \textcolor{blue}{blue} boxes and the ground truth is in \textcolor{red}{red}. LLM engine here is GPT-4 and the captioning model is LaViLa.}
\label{fig:viz_qa2}
\end{figure}

\looseness=-10000
We visualize the outputs of the LLM in~\Cref{fig:viz_nlq2} and~\Cref{fig:viz_qa2}. As shown by abundant qualitative examples, the LLM can produce high-quality answers in zero shot. It is worth noting that the machine-generated captions may contain objects that are not present in the video or miss critical information that can potentially answer the target query. Based on the imperfect captions, the LLM is still able to capture the key event and produce high-quality responses, suggesting a more powerful captioning model will further boost the performance. 

\clearpage

\clearpage
\section{Ambiguity in NLQ Annotations}
\label{ambiguity_in_nlq}

LLMs generate more than one interval when there are multiple temporal windows that can potentially answer the given query. We observe that some temporal windows proposed by the LLM seem reasonable despite only one ground-truth answer available in Ego4D NLQ annotations. These queries need to be filtered or modified to reduce ambiguity.

\begin{wrapfigure}{c}{0.9\textwidth}
\centering
\includegraphics[width=\textwidth,trim={0 0 1cm 0},clip]{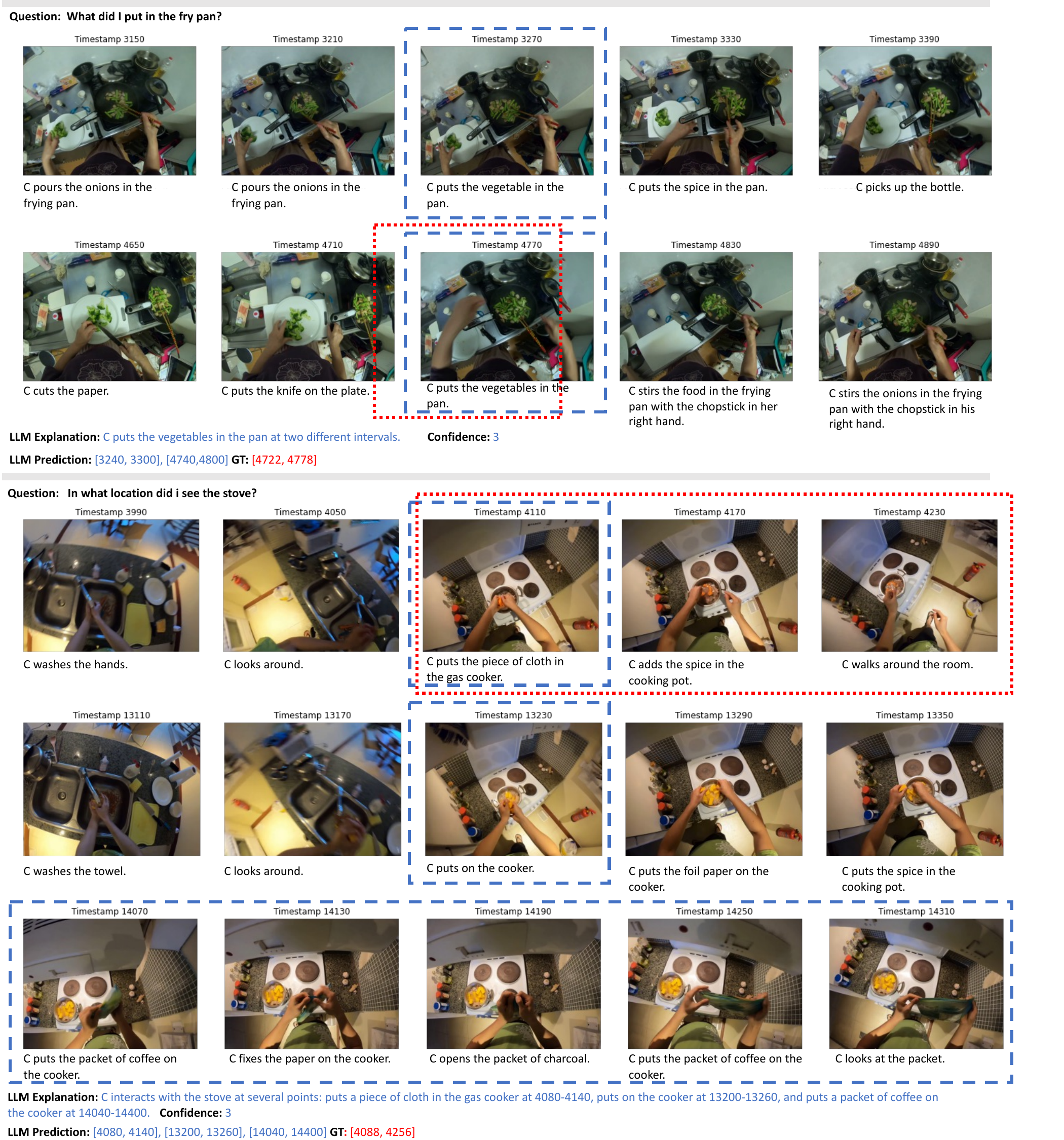}
\caption{Examples of Ambiguous Queries. LLM predictions are in \textcolor{blue}{blue} and the ground truth is in \textcolor{red}{red}.}
\label{fig:multi_preds}
\end{wrapfigure}

\clearpage

\end{document}